\documentclass[10pt,journal,compsoc]{IEEEtran}
\usepackage{times}
\usepackage{graphicx}
\usepackage{comment}
\usepackage{amsmath,amssymb} 
\usepackage{color}
\usepackage{epsfig}
\usepackage{dsfont}
\usepackage{calrsfs}
\usepackage{threeparttable}
\usepackage[linesnumbered, ruled]{algorithm2e}
\usepackage{algorithmicx}
\usepackage{mathtools}
\usepackage{array}
\usepackage{booktabs}
\usepackage{bm} 
\usepackage{amsfonts} 
\usepackage{multirow}
\usepackage{enumerate}
\usepackage{enumitem}
\usepackage{makecell}
\usepackage[subnum]{cases}
\usepackage{subcaption}
\usepackage{dsfont}
\usepackage{xspace}
\usepackage{diagbox}
\usepackage{soul, color}
\usepackage{balance}
\graphicspath{ {./images/} }
\usepackage[pagebackref=false,breaklinks=true,letterpaper=true,colorlinks,bookmarks=false]{hyperref}

\DeclareMathAlphabet{\pazocal}{OMS}{zplm}{m}{n}

\begin{document}

\title{Detecting Near-Duplicate Face Images}

\author{Sudipta Banerjee,~\IEEEmembership{Member,~IEEE,}
        and~Arun~Ross,~\IEEEmembership{Senior Member,~IEEE}
}


\maketitle

\begin{abstract}

Near-duplicate images are often generated when applying repeated photometric and geometric transformations that produce imperceptible variants of the original image. Consequently, a deluge of near-duplicates can be circulated online posing copyright infringement concerns. The concerns are more severe when biometric data is altered through such nuanced transformations. In this work, we address the challenge of near-duplicate detection in face images by, firstly, identifying the original image from  a set of near-duplicates and, secondly, deducing the relationship between the original image and the near-duplicates. We construct a tree-like structure, called an Image Phylogeny Tree (IPT) using a graph-theoretic approach to estimate the relationship, i.e., determine the sequence in which they have been generated. We further extend our method to create an ensemble of IPTs known as Image Phylogeny Forests (IPFs). We rigorously evaluate our method to demonstrate robustness across other modalities, unseen transformations by latest generative models and IPT configurations, thereby significantly advancing the state-of-the-art performance by $\sim 42\%$ on IPF reconstruction accuracy. Our code is publicly available at \url{https://github.com/sudban3089/DetectingNear-Duplicates}.

\end{abstract}

\begin{IEEEkeywords}
Image phylogeny, Near-duplicates, Convolutional graph neural network, Sensor pattern noise
\end{IEEEkeywords}

\IEEEpeerreviewmaketitle

\section{Introduction}

\begin{figure}[h]
     \centering
     \begin{subfigure}[b]{0.37\textwidth}
         \centering
         \includegraphics[width=\textwidth]{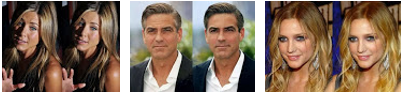}
        \end{subfigure} 
     \begin{subfigure}[b]{0.37\textwidth}
         \centering
         \includegraphics[width=\textwidth]{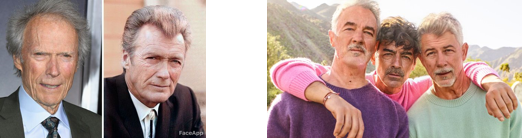}  
     \end{subfigure} 
     \begin{subfigure}[b]{0.37\textwidth}
         \centering
         \includegraphics[width=\textwidth]{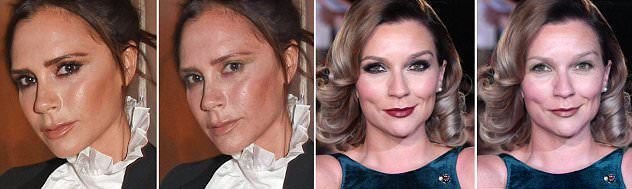}
        \end{subfigure} 
    
    \caption{Examples of near-duplicates available online illustrating celebrity face images modified through  (top row) Photoshop, (middle row) FaceApp and (bottom row) MakeApp filters. Image courtesy~\protect\cite{imagecourtesy}.}
\label{Fig:IntroEg}
 \end{figure}
Image editing tools can seamlessly generate near-duplicates by sequentially applying transformations to an original image. Photos, especially face images, appearing on social networking sites can be subjected to image \textit{enhancement} techniques for enhancing visual appeal~\cite{ImageEdit_1, ImageEdit_2}. Celebrity magazines sometimes use `airbrushing' to glamorize advertisements and this has drawn the attention of lawmakers.\footnote{\url{https://www.telegraph.co.uk/news/2017/09/30/photoshopped-images-come-warning-new-french-law/}} Photo-editing filters and generative models such as generative adversarial networks (GANs) and diffusion models can be used to alter the appearance of \textit{any} individual by modifying facial and demographic attributes such as age and hairstyle~\cite{BeautyGlow, AttGAN}. See Fig.~\ref{Fig:IntroEg}. 

\begin{figure}
\centering
\subfloat[]
{
    \includegraphics[scale=.93]{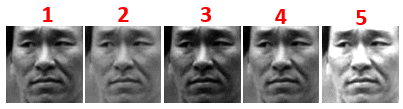} 
    } \\
 \subfloat[]
{
    \includegraphics[scale=.93]{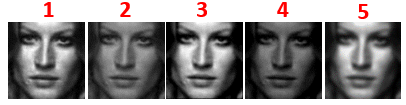} 
    }  
\caption{Two examples of near-duplicates generated using a sequence of photometric transformations \textemdash Brightness adjustment, Gamma transformation, Gaussian smoothing and Median filtering. In the top row, the original image is `2', and in the bottom row, the original image is `3'. Visual inspection cannot trivially determine the sequence in which the images were modified.}
\label{Fig:ImagePhoto}
\end{figure}

\begin{figure}
\centering
\subfloat[]
{
    \includegraphics[scale=.68]{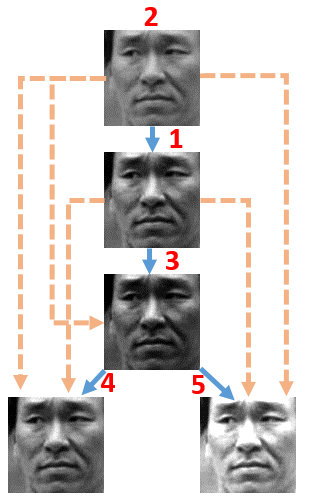} 
    } \hfill
 \subfloat[]
{
    \includegraphics[scale=.68]{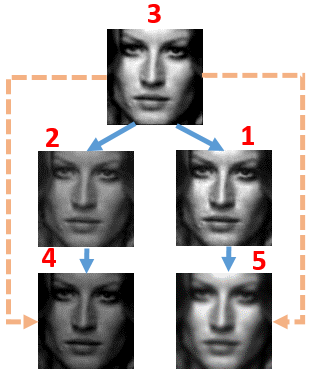} 
    }  
\caption{Two examples of image phylogeny trees (IPTs) corresponding to the two sets of near-duplicate images presented in Fig.~\ref{Fig:ImagePhoto}. The bold arrows indicate immediate links and the dashed arrows indicate ancestral links.}
\label{Fig:IPTs}
\end{figure}

{\textbf{Problem Definition.} Multiple digital modifications can result in an explosive number of near-duplicates that end up online with minimal regulation. In the wake of such unmitigated data duplication, provenance analysis has piqued the interest of academic and government entities\mbox{~\cite{ImageForensics_MediFor}}. It aids in tracing the evolution of the edited images and plays an integral role in media forensics. An image phylogeny tree (IPT) provides a visual assessment of the evolution and relationship between near-duplicates (see Figs.\mbox{~\ref{Fig:ImagePhoto}} and\mbox{~\ref{Fig:IPTs}}) and is used for their detection\mbox{~\cite{Barni_10_Depend, Dias_12_MST}}. IPT reconstruction begins with a set of near-duplicate images as input and generates a hierarchical structure as output. It consists of a root node (the original image) and directed edges (the relationship between the near-duplicates); the edges are directed from the parent node (source image) to the child node (transformed image). Although IPT consists of a single root node, there can be a situation involving multiple root nodes. An example can be a face image of an individual in the same scene but with different poses, expressions or captured using different cameras. In this case, transforming each image repeatedly, but independently of the others, will result in multiple IPTs. Each original image will then serve as a separate root, and each of them will span a distinct IPT. A collection of such IPTs will constitute an Image Phylogeny Forest (IPF)\mbox{~\cite{Dias_13_AutoIPF}}}. 

\begin{figure}
\centering
    \includegraphics[scale=.25]{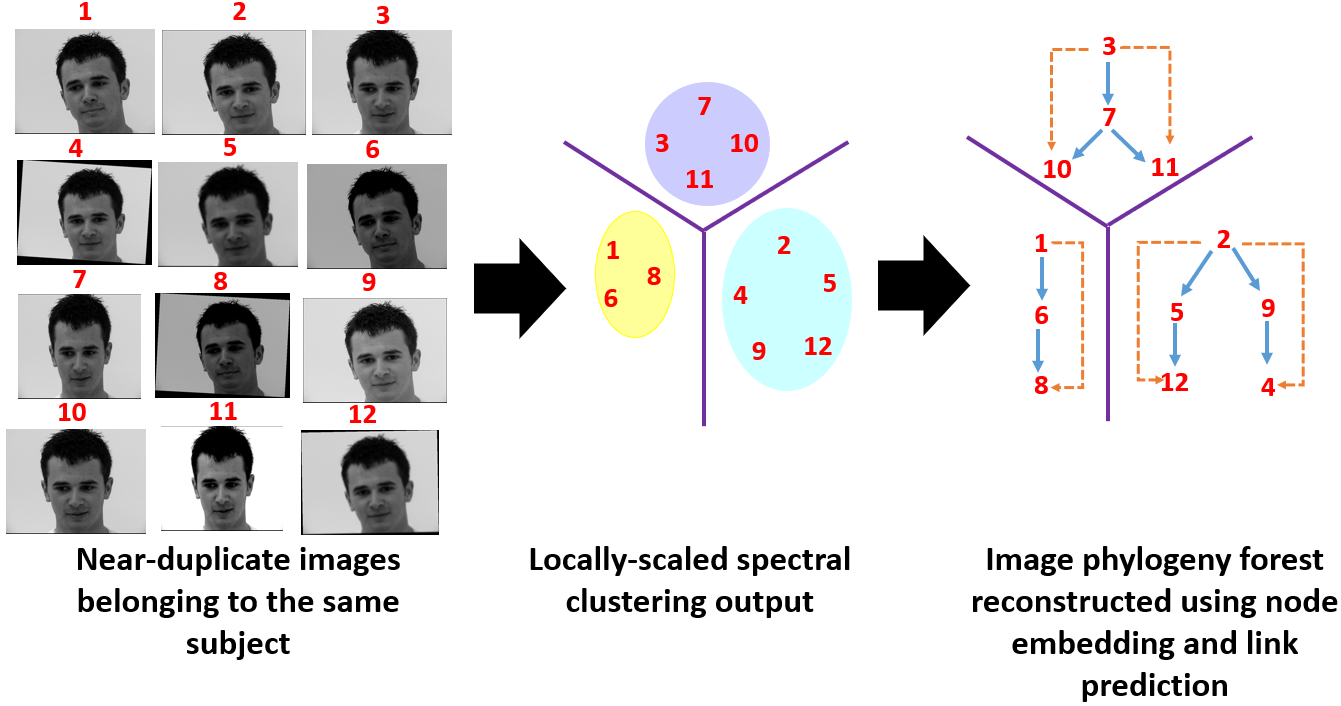} 
     
\caption{Outline of the objective in this work. Given a set of near-duplicate face images belonging to the same subject, our objective is two-fold. Firstly, we use locally-scaled spectral clustering to identify each cluster (each cluster represents an IPT). The clusters indicated by ellipsoids vary in diameter indicating the importance of local scaling. Secondly, we reconstruct each IPT using graph theoretic node embedding and link prediction. The ensemble of IPTs results in an Image Phylogeny Forest (IPF).}
\label{Fig:Objective}
\end{figure}

\textbf{Our Method:} Existing methods~\cite{Barni_10_Depend, Dias_13_largescaleIPT, Dias_13_OB, SpecClus_16_Dias, Bharati_17_UPhylogeny, Bharati_20_TE, Bestagini_20_ICASSP} focus on images depicting natural scenes and generic objects but face images pose a difficult challenge due to their inherent diversity and heterogeneity in terms of age, gender, ethnicity, expression, illumination, pose, and other attributes. Additionally, face images possess biometric utility that should not be hampered during processing. Therefore, we formulate the phylogeny tree construction as a graph-based framework where each image acts as a node, and the directed edges indicate the relationship between the nodes. A graph-based approach is well-suited for image phylogeny as it allows both pairwise and global analyses among the nodes. The nodes (images) are represented via their respective features, while the edges (interactions) are represented through an adjacency matrix that captures the relationship between them. We begin with an initial adjacency matrix and iteratively refine it until we arrive at the desired configuration resulting in the IPT. We employ a convolutional graph neural network that utilizes the feature values of the nodes to learn the interactions that best capture the relationship between the nodes. Further, we consider multiple IPTs and devise a simple technique to construct the image phylogeny forest (IPF) with high degrees of accuracy. See Fig.~\ref{Fig:Objective}.
The \textbf{contributions} of this work are as follows.

\noindent 1. We develop a technique that combines a graph neural network with sensor pattern noise (SPN) features to examine local and global interaction between the nodes, \textit{i.e.}, a set of near-duplicates. The graph network serves as a \textit{node embedding} module that identifies the hierarchical position of each near-duplicate node in an IPT. The SPN features perform \textit{link prediction} between each pair of nodes embedded by the graph network. The node embedding and link prediction modules work in tandem to perform complementary local and global analyses to generate the IPT. We extend our method to create an IPF using a locally-scaled spectral clustering method that identifies disjoint IPTs. \\
\noindent 2. We demonstrate that our method is domain agnostic through empirical evaluations. Leveraging the graph theoretic approach offers the flexibility to be applied to unseen transformations, unseen IPT configurations, across different biometric modalities, and even to other natural scene images.\\ 
\noindent 3. We perform a rigorous analysis of the proposed method across different datasets representing real-world scenarios and compare it with baselines demonstrating significant improvement. Finally, we release the codes and data for reproducible research.

\section{Related Work}
\label{RelWork}
Near-duplicate detection and retrieval is a well-studied problem in the literature~\cite{NDDR_Ref1, NDDR_Ref2, NDDR_Ref3}. Finding relationship between near-duplicates, \textit{i.e.}, determining the sequence of evolution between the near-duplicate images, is studied in~\cite{ ImageForensics_1,Barni_10_Depend,Dias_13_largescaleIPT,Moreira_18_Provenance,Bharati_17_UPhylogeny,Bharati_2}. The relationship between a set of near-duplicate images can be represented using an \textit{Image Phylogeny Tree}~\cite{Dias_12_MST,Dias_13_OB,Melloni_14_dissmetricsIPT,Oliveira_14_multipleparentingIPT}. Conventionally, IPT construction involves two steps~\cite{FirstIPT,Dias_12_MST}: \\
\noindent (i) Computing an asymmetric (dis)similarity measure between every pair of images in the set. (ii) Using a tree spanning algorithm to infer the links between image pairs and identify the parent (original) and child (transformed) nodes based on the asymmetric measure.
The first step computes an asymmetric measure that models the relationship between each pair of near-duplicate images in the set. Typically, a majority of the existing methods~\cite{Dias_12_MST,Melloni_14_dissmetricsIPT,Moreira_18_Provenance} perform pairwise modeling to compute the asymmetric measure. The objective is to minimize some distance or error function between the pair of images. However, while performing the pairwise modeling, these methods \textit{do not} consider the global relationship between the images. 

`Global' relationships were explored by utilizing all the images simultaneously in~\cite{Bestagini_20_ICASSP}, where the authors use a de-noising autoencoder to improve the dissimilarity matrix to obtain an accurate IPT. The authors in~\cite{Bharati_20_TE} used a deep neural network to rank the order in which the near-duplicates have been generated by learning the transformation-specific embedding. The majority of these methods treat IPT as a minimal spanning tree, \textit{i.e.}, each node (image) has a single incoming edge. Some methods~\cite{Ban_17_IJCB,Ross_19,TBIOM_20} relaxed the minimal spanning tree constraint to include ancestral links, making the IPT a directed acyclic graph (DAG), as implemented in~\cite{Moreira_18_Provenance} to create provenance graphs. {Other work include discovering image manipulation history through forensic analysis, feature-agnostic large-scale image retrieval and provenance evaluation of large-scale datasets~\mbox{\cite{NewRef_1, CBIR_2021, Moreira2022, MFCProv}}}.

Image phylogeny forest reconstruction typically involves two forms of approaches. (i) Consider IPF reconstruction as an extension of the IPT reconstruction process. Initially, each node is considered as an individual IPT, and then they are successively merged until a terminating criterion is met~\cite{Dias_13_AutoIPF,Costa_14_IPF,Oliveira_16_IPF}. (ii) Consider IPF reconstruction as a two-step process, where the first step clusters the images, (each cluster represents an IPT), and the second step constructs the IPT corresponding to each cluster~\cite{SpecClus_16_Dias}. 


In real-world applications, we have no prior knowledge about the number of IPTs, or the number of nodes within each IPT. Pairwise modeling of relationship in an IPT provides limited analysis and impairs accurate IPT reconstruction. Therefore, we propose a novel IPF reconstruction method with two components: (ii) A graph-based approach coupled with sensor pattern noise features to model both local (pairwise) and global relationship to accurately reconstruct the IPT, and (ii) An improved spectral clustering algorithm that uses locally-scaled kernels to handle varying number of IPTs to create an IPF. \textbf{The novelty of the proposed method lies in framing image phylogeny as a graph-theoretic problem, in which, the near-duplicate images serve as nodes and the sequence of transformations serve as the edges between them. The idea is to determine the `best' graph structure that will capture the relationship between the near-duplicates in the form of an image phylogeny tree.} The re-formulation of the problem allows domain-agnostic application. Therefore, the proposed method can be applied to different modalities (natural scene images and biometric images) and different transformations (manual editing and deep-learning based manipulations). The remainder of the paper is organized as follows. Sec.~\ref{Prop} describes the proposed method used for image phylogeny. Sec.~\ref{Imp} outlines the implementation details adopted in this work. Sec.~\ref{DataandExpts} describes the dataset and the experimental protocols used in this work. Sec.~\ref{Res} presents the results. Sec.~\ref{Concl} concludes the paper with future work. 

\section{Proposed Method}
\label{Prop}
The proposed IPF reconstruction method has been broadly outlined in Fig.~\ref{Fig:Objective}. The proposed method has two phases or steps \textemdash \textbf{grouping} and \textbf{phylogeny}. {Grouping involves clustering the input near-duplicates into disjoint groups, where each group will correspond to a single IPT. We initially explored deep-learning based clustering methods. Deep embedding for clustering analysis\mbox{~\cite{DEC}} allows joint optimization of the best set of features from the input and then categorizing them into the best possible groups or clusters. This is a form of unsupervised learning of features and cluster memberships without ground-truth labels. The method outperforms traditional algorithms on image and text datasets. Note that face images have diverse structural and textural details and are evidently more difficult to cluster compared to generic images. Therefore, we focus on classical techniques that might be better adapted to the current task.  Clustering algorithms\mbox{~\cite{Cluster_Jain}} such as k-means\mbox{~\cite{kmeans}}, density based clustering\mbox{~\cite{dbscan}} and spectral clustering\mbox{~\cite{specclus}} exist. k-means requires the number of clusters to be specified, and is sensitive to noise and outliers. Density-based clustering methods although effective for arbitrary shaped clusters, are sensitive to parameter selection. Spectral clustering conceptualizes data points as nodes in a graph and interprets the distances between data points as walks on the graph lying on a non-linear manifold. Spectral clustering suffers from a major limitation that it is incapable of handling multi-scale cases as discussed in~\mbox{\cite{Geom_Spec_15}}. The multi-scale limitation posed by conventional spectral clustering concerns our work as we do not assume the scale or the size of the cluster apriori, \textit{i.e.}, the number of nodes (images) in each IPT. Therefore, in this work we use \textit{locally-scaled spectral clustering} for grouping the images to distinct IPTs.} See the details in Sec.~\ref{Sec:LS-SC}.

Next, we proceed to the second phase of our IPF pipeline, \textit{i.e.}, recovering the phylogeny structure within each cluster. Our method couples graph theory with sensor pattern noise features to explore macroscopic (global) and microscopic (pairwise) analysis of the images. Consider the near-duplicates as nodes or vertices arranged in a high-dimensional space such that they are related to each other because they are derived from each other in a specific order. The order in which they have been modified can then be represented as a `walk' from one node to the other. IPT reconstruction can therefore be considered as determining the correct sequence of walks between the nodes which will indicate the global relationship between a set of near-duplicates. We use \textit{node embedding} to derive this global relationship. Node embedding converts a high-dimensional input (an image) to a low-dimensional output (a scalar value). The scalar value generated by the node embedding should indicate the global relationship of a node. But the challenge is how can a single scalar value represent a global property? We formulated the problem as follows: we will use node embedding to deduce the depth label (a scalar value) of a node (image). The \textit{depth label} of an image (say, $k$) in an IPT signifies the number of ancestors ($k-1$) for that image, with respect to the root node located at depth=1. For example, an image at depth=2 has only 1 ancestor, \textit{i.e.}, the root node, while an image at depth=3 has 2 ancestors. The depth labels of the images in an IPT provide how the images are globally related with each other. However, the depth labels cannot determine the relationship between the nodes at successive depths. For example, there can be multiple nodes with depth labels $s$ but only one of them will be the immediate parent of node associated with depth label $t$ (where, $t= s+1$). Therefore, we use a \textit{link prediction} module to identify the existence of links between nodes located at depth $k$ and depths $>k$ by performing a pairwise microscopic analysis. See details of the node embedding module in Sec.~\ref{NE} and the link prediction module in Sec.~\ref{IPT}.

\subsection{Locally-scaled spectral clustering}
\label{Sec:LS-SC}
Spectral clustering comprises the following steps.\\
\noindent 1. Construct a symmetric similarity matrix $\bm{S}$ and its corresponding adjacency matrix $\bm{A}$ from $Q$ data points $\bm{X}$. \\
\noindent 2. Compute the graph Laplacian matrix $\bm{L}$ from $\bm{A}$.\\
\noindent 3. Perform eigen decomposition on $\bm{L}$ and select $k$ eigen vectors $\bm{U}$ corresponding to $k$ smallest eigen values. \\
\noindent 4. Apply k-means on the rows of the $\bm{U}$ to obtain cluster assignment for each data point.

The notion of scale arises in the similarity matrix $\bm{S}$. It uses a kernel, $h$ (preferably a smooth kernel such as an exponential decay function) to compute the affinity between a pair of points $(\bm{X}_i, \bm{X}_j)$ as $\bm{S}(i,j)= h\bigg(\frac{\| \bm{X}_i - \bm{X}_j \|^2}{\sigma^2} \bigg)$. The parameter $\sigma$ refers to the bandwidth variable, and is usually computed by training on a large number of data points. An accurate $\sigma$ is pivotal but is governed by the implicit assumption that all the clusters have similar scales, \textit{i.e.}, approximate uniform distribution of data points across all the clusters. A single global bandwidth may result in incorrect merging of disjoint data groups. Alternatively, \textbf{locally-scaled} kernels vary with the data points resulting in invariance to sampling density~\cite{Geom_Spec_15}. Then the similarity matrix can be formulated as $\bm{S}(i,j)= h\bigg(\frac{\| \bm{X}_i - \bm{X}_j \|^2}{\sigma(\bm{X}_i) \sigma(\bm{X}_j)} \bigg)$. The variables $\sigma(\bm{X}_i)$ and $\sigma(\bm{X}_j)$ are known as local bandwidths.
 
We apply the local-scaling on the feature space ($f(\bm{X})$) instead of directly applying it on the data space ($\bm{X}$) to allow aggregation of multiple features for accurate clustering. We selected three features in this work: (i) pixel intensity, (ii) sensor pattern noise features, particularly Enhanced Photo Response Non-Uniformity (PRNU)~\cite{Li_TIFS_10}, and (iii) face descriptors. Pixel intensity and PRNU features account for the variation in the acquisition settings and devices. Face descriptors\footnote{We can always substitute face descriptors  with generic image descriptors for images depicting natural scenes.} account for variation in pose and expression. We use VGGFace descriptors~\cite{VGGFace} in this work. The resultant similarity matrix now becomes, 
\begin{equation*}
\bm{S}(i,j) = \exp - \bigg(\frac{\bm{D}_F(i,j)^2}{\hat{p}_F(i)\hat{p}_F(j)} + \frac{\bm{D}_N(i,j)^2}{\hat{p}_N(i)\hat{p}_N(j)} + \frac{\bm{D}_P(i,j)^2}{\hat{p}_P(i)\hat{p}_P(j)} \bigg).
\end{equation*} 

{The cosine similarity between face descriptors (vectors) computed from a pair of face images $(i,j)$, \textit{i.e.}, $\textbf{F}_i$ and $\textbf{F}_j$ is $cosinesim(\textbf{F}_i, \textbf{F}_j) = \frac{\textbf{F}_i^T \textbf{F}_j}{\lVert\textbf{F}_i\rVert \lVert\textbf{F}_j\rVert}.$ Considering the cosine similarity to vary between $[-1,1]$, we first normalize it to $[0,1]$ via $0.5 \times (cosinesim+1)$. Then we use the normalized value to compute a distance measure as $\textbf{D}_F(i,j)= 1- \frac{\textbf{F}_i^T \textbf{F}_j}{\lVert\textbf{F}_i\rVert \lVert\textbf{F}_j\rVert}.$ Similarly, cosine distance between pixel features, $P(\cdot)$, can be expressed as $\textbf{D}_P(i,j)= 1- \frac{\textbf{P}_i^T \textbf{P}_j}{\lVert\textbf{P}_i\rVert \lVert\textbf{P}_j\rVert}.$ Finally, cosine distance between PRNU-based sensor noise features, $N(\cdot)$, can be expressed as
$\textbf{D}_N(i,j)= 1- \frac{\textbf{N}_i^T \textbf{N}_j}{\lVert\textbf{N}_i\rVert \lVert\textbf{N}_j\rVert}.$ 

 The terms in the denominator are locally-scaled bandwidths computed using univariate kernel density estimate of the respective features. $\hat{p}_F(\cdot)$ corresponds to kernel density estimate (KDE) computed from face descriptors, $\hat{p}_F(\cdot)=\frac{1}{nb}\sum_{i=1}^n h \bigg(\frac{F(\cdot) - F(x_i)}{b}\bigg)$, where $n$ is the number of data points, $b$ is the bandwidth equal to the interval between all the points in $x$ and $h$ is a normal kernel. Similarly, $\hat{p}_N(\cdot)$ denotes KDE computed from the PRNU features, and $\hat{p}_P(\cdot)$ denotes KDE computed from the pixel intensity features. Next, we compute binarized similarity matrix $\bm{S}_B$ using median of each row as threshold, such that $\displaystyle \bm{S}_B(i,j)=1$ if $\bm{S}(i,j)>median(\bm{S}(i,:))$, and $\bm{S}_B(i,j)=0$, otherwise. Then we follow the steps pertaining to conventional spectral clustering, such as computation of the degree matrix, Laplacian matrix, and eigen decomposition. The diagonal degree matrix for $Q$ nodes is computed as $\displaystyle \bm{Deg}(i,i) = \sum_{j=1}^Q\bm{S}_B(i,j)$. The normalized Laplacian is computed as $\displaystyle \bm{L} = \bm{Deg}^{-\frac{1}{2}}\bm{S}_B\bm{Deg}^{-\frac{1}{2}}$.} Eigen value decomposition of the Laplacian matrix results in eigen vectors ($\bm{U}$) and eigen values ($\Lambda$). We apply a threshold $\eta$ (selected using a validation set) to obtain a subset of eigen values, say $k$ smallest eigen values such that, $\{\Lambda_k \subset \Lambda | \Lambda_k < \eta \}$. The corresponding eigen vectors become $\bm{U}_k = [\bm{u}_1, \bm{u}_2, \cdots, \bm{u}_k]$. Finally, \textit{k}-means is applied to the rows of $\bm{U}_k$ to get cluster assignments, $\mathcal{C} \in \mathbb{R}^{Q \times 1}$ for each image, such that $\displaystyle \mathcal{C} =$ k-means$(\bm{U}_k, k)$, where, $k$ is the number of clusters. See Algo.~\ref{Alg:LC-SC}.

\begin{algorithm}
\footnotesize
  \caption{\label{Alg:LC-SC}Locally-scaled spectral clustering}
  \KwIn{Set of $Q$ near-duplicate images $\bm{I}$}
  \KwOut{Cluster assignments $\mathcal{C}$}
  
  Compute the face descriptors $\bm{F}$ from images (we used VGGFace)
  
  Compute the PRNU $\bm{N}$ from images (we used Enhanced PRNU)
  
  Compute the vectorized representation of the pixel intensities $\bm{P}$ from images
  
  Compute kernel density estimate for face descriptors, PRNU features and pixel features, respectively: $\hat{p}_F(\cdot), \hat{p}_N(\cdot), \hat{p}_P(\cdot)$
  
  \Repeat{$i, j \leq Q$}{
   Compute distance values between each pair of images ($\bm{I}_i, \bm{I}_j$) for face descriptors, PRNU features and pixel features, respectively (we used cosine distance) \hspace{5cm}
  {$\textbf{D}_F(i,j)= 1- \frac{\textbf{F}_i^T \textbf{F}_j}{\lVert\textbf{F}_i\rVert \lVert\textbf{F}_j\rVert}$ \\ 
   $\textbf{D}_N(i,j)= 1- \frac{\textbf{N}_i^T \textbf{N}_j}{\lVert\textbf{N}_i\rVert \lVert\textbf{N}_j\rVert}$  \\
  $\textbf{D}_P(i,j)= 1- \frac{\textbf{P}_i^T \textbf{P}_j}{\lVert\textbf{P}_i\rVert \lVert\textbf{P}_j\rVert}$}  \hspace{5cm}  
  \nl Compute symmetric similarity matrix $\bm{S}$: \hspace{5cm}
  $\bm{S}(i,j) = \exp - \bigg(\frac{\bm{D}_F(i,j)^2}{\hat{p}_F(i)\hat{p}_F(j)} + \frac{\bm{D}_N(i,j)^2}{\hat{p}_N(i)\hat{p}_N(j)} + \frac{\bm{D}_P(i,j)^2}{\hat{p}_P(i)\hat{p}_P(j)} \bigg)$ 
}

Compute binarized similarity matrix $\bm{S}_B$ using median of each row as threshold

Compute diagonal degree matrix $\bm{Deg}$

Compute normalized Laplacian $\bm{L}$

Perform eigen value decomposition to obtain eigen vectors ($\bm{U}$) and eigen values ($\Lambda$) from Laplacian

Select thereshold $\eta$ to select $k$ smallest eigen vectors $\Lambda_k$: $\bm{U}_k = [\bm{u}_1, \bm{u}_2, \cdots, \bm{u}_k]$

Apply \textit{k}-means clustering on the rows of $\bm{U}_k$ to get cluster assignments $\mathcal{C}$ for each image

Return $\mathcal{C}$
\end{algorithm}

\begin{figure*}[t]
 \centering
    \includegraphics[scale=.4]{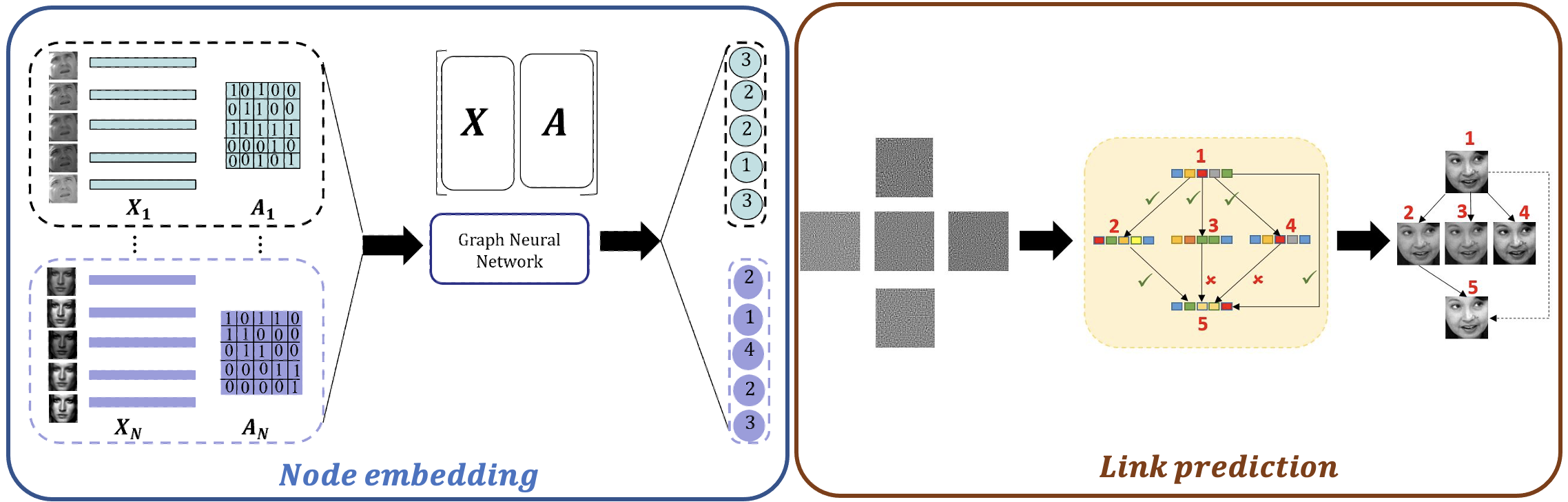} 
\caption{{Outline of the two-step method used in this work. In the first step, the \textit{node embedding} module, i.e., a graph neural network, $f\bm{(X,A)}$ accepts $\bm{X}$: features from each image in the IPT; features can be pixel intensity or convolutional features, and $\bm{A}$: an initial adjacency matrix as a pair of inputs. During \textit{training}, the initial adjacency matrix denotes the ground truth relationship between the images in the IPT, while during \textit{testing}, the initial adjacency matrix is populated by comparing distance between images features against a threshold to determine presence of edge $(=1)$ or absence of edge $(=0)$. The node embedding module produces a vector of depth labels corresponding to each IPT configuration. In the second step, the \textit{link prediction} module accepts $g\bm{(l,N)}$, $\bm{l}$: depth labels from the first module, and $\bm{N}$: sensor pattern noise (PRNU) features computed from the near-duplicates as inputs. The link prediction module produces the reconstructed image phylogeny tree (IPT).}}
 \label{Fig:2stepmethod}
\end{figure*}

\subsection{Node Embedding}
\label{NE} 

The node embedding module accepts a set of related near-duplicate images $\bm{I} \in \mathbb{R}^{M \times d \times d}$ and outputs a vector of depth labels $\bm{l} \in \mathbb{R}^{M}$. Here, $M$ refers to the number of near-duplicate images belonging to an IPT or cluster, where $M < Q$, (total number of images), and each image is of size $d \times d$. Let $f(\cdot, \cdot)$ represent the node embedding module which requires a pair of inputs $\bm{X}$ and $\bm{A}$. Therefore, $\bm{l}=f(\bm{X},\bm{A})$. Here, $\bm{X}$ refers to image features which can be pixel intensity values or convolutional feature maps (see Sec.~\ref{Sec:Ablation}), and $\bm{X}\in \mathbb{R}^{M \times d^2}$. $\bm{A}$ represents the adjacency matrix of size $M \times M$ that represents the relationship between the images. The adjacency matrix can be computed using a nearest-neighbor based method utilizing features extracted from the images. Note that the adjacency matrix does not necessarily encode the IPT structure. See Fig.~\ref{Fig:2stepmethod}~\textit{Node embedding} module.

\subsection{Link Prediction}
\label{IPT}
The link prediction module accepts a vector of depth labels $\bm{l}\in \mathbb{R}^M$ for a set of $M$ near-duplicate images and outputs a data structure containing a set of vertices $V$, such that $|V|= M$, and directed edges $E$. We refer this directed structure as the \textit{image phylogeny tree}: $\bm{IPT}(V,E)$. Let $g(\cdot, \cdot)$ represent the link prediction module that requires a pair of inputs $\bm{l}$ and $\bm{N}$. Therefore, $\bm{IPT} = g(\bm{l},\bm{N})$. Here, $\bm{l}$ refers to the depth labels computed using the node embedding step. $\bm{N}$ represents the sensor pattern noise features computed from the images $\bm{I}$, such that, $\bm{N} \in \mathbb{R}^{M\times d^2}$. The link prediction performs two tasks: (i) depth label correction, and (ii) pairwise link inference.

\begin{algorithm}
\footnotesize
  \caption{\label{Alg:IPTCon}Link prediction}
  \KwIn{Set of $M$ near-duplicate images $\bm{I}$, set of depth labels $\bm{l}$ provided by GNN, where $|\bm{l}|=M$ }
  \KwOut{Root $R$ and $\bm{IPT}$}
  
   Extract the sensor pattern noise (PRNU) for each image $\displaystyle \bm{N}(i)$ where $i=1,\cdots,M$
 
  Identify whether multiple nodes have depth label = 1, if $|Candidateroots| > 1$ go to Step 3, else go to Step 4  
  
  Perform depth label correction
  
  Infer pairwise links
   
  Identify the root node as the node with corrected depth label = 1, $R$
   
  Construct the phylogeny tree: $\bm{IPT}(V,E)$ \hspace{3cm}
   \Comment $V$ represents the set of nodes and $E$ represents the set of directed edges
  
  Return $R$ and $\bm{IPT}$
     
\end{algorithm}

\begin{figure}
\centering
    \includegraphics[scale=.35]{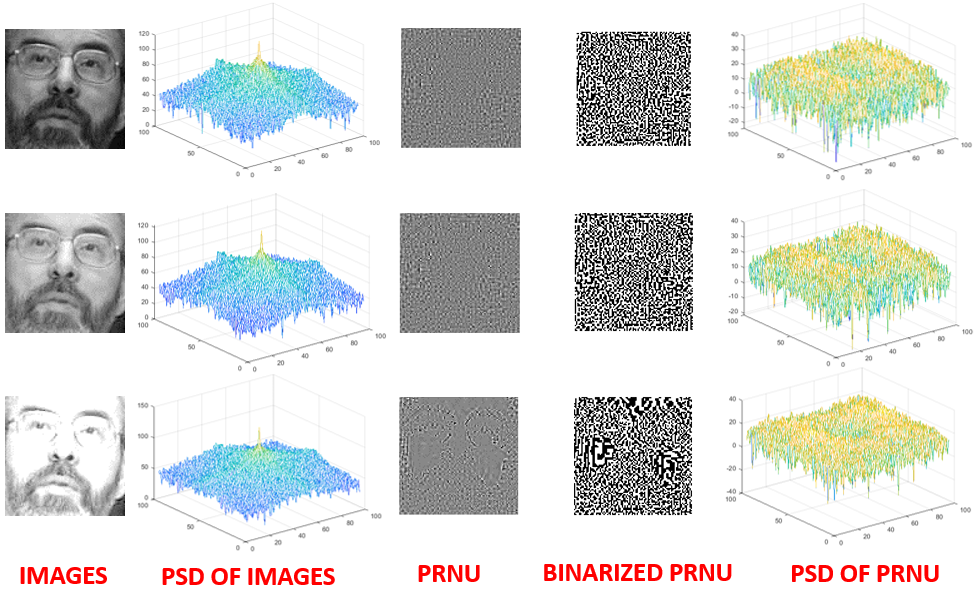} 
     
\caption{{Photo Response Non-Uniformity (PRNU) signatures for three near-duplicates are presented. The variations in the PRNU patterns (grayscale images) are better visualized using the binary maps (threshold=0) and their corresponding power spectral density (PSD) plots. Note that the PSD plots of the PRNU patterns reveal discernible differences compared to the images demonstrating that PRNU effectively distinguishes between near-duplicates.}}
\label{Fig:PRNUPhylo}
\end{figure}

For depth label correction, we first check whether multiple nodes have depth label=1, because an IPT can have only one root node. Multiple root nodes exist if $|Candidateroots| > 1$, where, $\displaystyle Candidateroots = arg_{i} \{\bm{l}(i)==1 \}$, where $i=1,\cdots,M$. We employed sensor pattern noise features present in images for depth label correction. Photo Response Non-Uniformity (PRNU) is a type of sensor pattern noise which manifests in the image as a result of non-uniform response of the pixels to the same light intensity~\cite{Lukas_TIFS_06}, and is typically used for sensor identification. Photometric and geometric transformations can induce changes in the PRNU pattern present in an image~\cite{Lukas_TIFS_06,Ross_18}. See Fig.~\ref{Fig:PRNUPhylo} to visualize how PRNU patterns change in presence of transformations. We used the power spectral density (PSD) plots to demonstrate variations in PRNU patterns that can help discriminate between original and transformed images. Firstly, we compute the Euclidean distance between the PRNU features of each of the candidate root nodes and the remaining nodes, $\bm{D}_c = \|(\bm{N}(Candidateroots(c))-\bm{N}(z))\|^2_2$, where $1 \leq z \leq M, z \neq c$, $\bm{N}$ corresponds to PRNU for each image. As the depth increases, it implies that the root node (original image) has undergone multiple sequences of transformations, resulting in higher variation in PRNU patterns. So, we retain that node which results in the highest distance as the \textit{correct} root node, $l(c) =1$, where, $c = argmax(\bm{D})$. Finally, we deduce the root node as the node with corrected depth label = 1, $R = \arg (\bm{l}_{corr}(i)==1), 1 \leq i \leq M$, where, $\bm{l}_{corr}$ represents the set of corrected depth labels. {The remaining nodes that were misclassified as depth label=1 are then re-assigned to depth label=2. The reason for re-assignment to depth label=2 is simply because higher depth labels would imply missing intermediate nodes. So, to avoid ambiguity we assign them to depth label=2.}

Pairwise link inference determines the existence of links between nodes located at depth labels $k$ and $>k$. Consider nodes located between successive depths, say, nodes $r$ and $s$ that are assigned the same depth label, $k$, by the node embedding module, and another node $t$ that is assigned depth label $k+1$. We need to identify the correct parent of node $t$ (see Fig.~\ref{Fig:2stepmethod}~\textit{Link prediction}). So, we compute the squared $L_2-$ norm between their PRNU features: $\displaystyle D_{rt} = \|(\bm{N}(r)-\bm{N}(t))\|^2_2$ and $\displaystyle D_{st} = \|(\bm{N}(s)-\bm{N}(t))\|^2_2$. We select that node as the parent of $t$ which results in the least distance,\footnote{Unrelated nodes or ancestors are more likely to result in higher Euclidean distance with respect to their PRNU features.} \textit{i.e.}, $r \rightarrow t$ ($r$ is the parent) if $ \displaystyle D_{rt} < D_{st}$; otherwise, $s \rightarrow t$ ($s$ is the parent). Refer to Algo.~\ref{Alg:IPTCon}.  

{\textbf{Feature Selection.}
Our method performs the first step of clustering or grouping to correctly identify the number of trees in the forest. Images captured by different sensors at different times should be assigned to different groups. After grouping, the node embedding step performs depth label prediction for the near-duplicates in \textit{each} group. The number of features needed depends on the complexity of the task. Both spectral clustering and node embedding steps analyze all images \textit{globally} that require multiple rich features such as face descriptors, pixel intensity values and sensor pattern noise. Finally, link prediction infers edge/link direction between a pair of nodes \textit{locally}. Convolutional features can be agnostic to photometric transformations, while pixel intensity is not sufficient to distinguish between parent (original) and child (transformed image). Hence, we use PRNU as it can effectively distinguish between parent and child nodes.}

\section{Implementation}
\label{Imp}
In this work, we explored three graph neural networks (GNNs),\textit{ viz.}, (i) ChebNet~\cite{GCN_NIPS_16}, (ii) Graph Convolutional Network (GCN)~\cite{GCN_Linear_17} and (iii) Hypergraph Neural Network (HGNN)~\cite{HGNN_19} to accomplish the task of node embedding. 
The implementation of the ChebNet and GCN is based on~\cite{GCN_Linear_17}.\footnote{\url{https://github.com/tkipf/gcn}} The parameters used are as follows: learning rate = 0.01, number of epochs = 100, number of units in hidden layer = 16, dropout = 0.5, weight decay = $5\times 10^{-4}$, early stopping = 10 (iterations) and degree of Chebyshev polynomial, $K = \{3,\cdots ,9\}$ (for ChebNet only). The cross-entropy loss function is employed. Both ChebNet and GCN require a feature matrix and an adjacency matrix. We use either the pixel intensity values or the PRNU features extracted from the images as the feature matrix of size $M \times D^2$, where, $M$ is the number of images (in each set) and each image from the set is of size $D \times D$. For extracting the PRNU features, we used the method described in~\cite{Li_TIFS_10}. The method accepts a block diagonal matrix as the adjacency matrix input; each block represents a single $M \times M$ adjacency matrix corresponding to one set of near-duplicates. During \textit{training}, we used the actual IPT configuration to construct the adjacency matrix. An asymmetric matrix can be used as adjacency matrix provided correct normalization is used for computing the graph Laplacian ($\bm{D}^{-1}\bm{A}$ instead of $\bm{D}^{-\frac{1}{2}}\bm{A}\bm{D}^{-\frac{1}{2}}$, where $\bm{D}$ is the degree matrix and $\bm{A}$ is the adjacency matrix). Note that this degree matrix and adjacency matrix (for each near-duplicate set) are \textit{different} from the ones used in locally-scaled spectral clustering (multiple near-duplicate sets). During \textit{testing}, we do not assume any prior IPT configuration or image transformation. During testing, first we compute the PRNU features, next we compute the squared $L_2-$norm between them, and assign a link between the node pair if the distance is less than some threshold (\textit{e.g.}, mean). 

For HGNN,\footnote{\url{https://github.com/iMoonLab/HGNN}} the parameters used are as follows: learning rate = 0.001, number of epochs = 600, number of units in hidden layer = 128, dropout = 0.5, weight decay = $5\times 10^{-4}$, multi-step learning rate scheduler parameters: gamma = 0.9, decay step = 200, decay rate = 0.7, milestones = 100 and the cross-entropy loss function. We used pixel features and PRNU features separately but we observed best results when both features are used together. The hypergraph adjacency matrix, $\in \mathbb{R}^{M \times 2M}$, is constructed by concatenating horizontally the adjacency matrices corresponding to the pixel features and PRNU features. Each adjacency matrix is constructed using the $k-$nearest neighbor ($k=5$) method. Our implementation is available at.\footnote{\url{https://github.com/sudban3089/DetectingNear-Duplicates}} 

\begin{table}[t]
\centering
\caption{Photometric and geometric transformations and the range of the corresponding parameters used in Experiments 2 and 3. The transformed images are scaled to $[0,255]$. Note that these transformations are being used only in the training stage. For the test stage, any arbitrary transformation can be used. }
\label{Tab:Params}
\scalebox{0.78}{
\begin{tabular}{|l|l|l|l|}
\hline
\textbf{Transformations} & \textbf{\begin{tabular}[c]{@{}l@{}}Level of \\ Operation\end{tabular}} & \textbf{Parameters}               & \textbf{Range}                                                      \\  \hline \hline
Brightness adjustment                & Global                                                                 & [a,b]                             & a $\in$ [0.9,1.5], b $\in$ [-30,30]                                            \\
Median filtering                     & Local                                                                  & size of window [m,n]              & m $\in$ [2,6], n $\in$ [2,6]                                              \\
Gaussian smoothing                   & Global                                                                 & standard deviation                & $\sigma \in$ [1,3]                                                      \\
Gamma transformation                     & Global                                                                 & gamma                             & $\gamma$ $\in$ [0.5,1.5]  \\ \hline                                                  \hline
Translation                     & Global & [$T_x$, $T_y$]              & $T_x \in$ [5,20], $T_y \in$ [5,20]                                              \\
Scaling                  & Global                                                                 & Percentage               & [90\%, 110\%]\\
Rotation                   & Global                                                                 & theta                             & $\theta \in [-5^0,5^0]$  \\ \hline                                                 

\end{tabular}}
\end{table}
\section{Datasets and Experiments}
\label{DataandExpts}

\subsection{Expt. 1: Evaluation of locally-scaled clustering}
\label{Expt1}
In this experiment, we used near-duplicates 1) downloaded from the Internet, and 2) generated using deep learning-based transformations. We used Google search query such as \textit{Angelina Jolie} and \textit{Superman} to download near duplicates from the \textbf{Internet} (see Fig.~\ref{Fig:SC_Internet}). These images are acquired in the wild without sensor information. We also used 22 images belonging to four subjects from~\cite{AttGAN}, where \textbf{deep learning}-based manipulations are applied on images to alter attributes such as adding hair bangs or adding glasses, generating near-duplicates (see Fig.~\ref{Fig:SC_GAN}). We applied locally-scaled spectral clustering on these images to discover how many IPTs are present in an IPF. Note in this experiment, we do not know about the sensor information. It can be presumed that the images curated from the Internet may have been acquired using different sensors. Therefore, this experiment represents ``in-the-wild" evaluation.

\subsection{Expt. 2: IPT reconstruction}
\label{Expt2}
\begin{figure*}[h]
\centering

    \includegraphics[scale=.45]{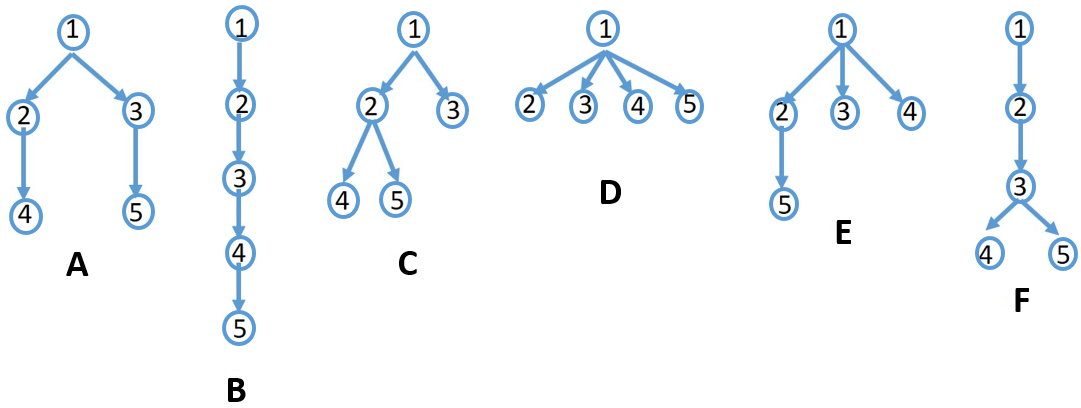}

\caption{IPT configurations (structures) used in Experiment 2. For ease of visualization, only the immediate links are depicted. However, the ancestral links are also included for evaluation.}
\label{Fig:Expt2_IPT}
\end{figure*}

In this experiment, we evaluated the proposed method in terms of (i) root identification accuracy, which computes the proportion of correctly identified root nodes, and (ii) IPT reconstruction accuracy, which is computed as follows: 
$\displaystyle \frac{|\textit{Original\textunderscore edges} \cap \textit{Reconstructed\textunderscore edges}|}{|\textit{Original\textunderscore  edges}|}$ for face images subjected to four \textbf{photometric} (Brightness adjustment, Gamma transformation, Median filtering and Gaussian smoothing) and three \textbf{geometric} (Rotation, Scaling and Translation) transformations. We used face images from the Labeled Faces in the Wild (LFW) dataset~\cite{LFWTech}. All the images are cropped to a fixed size of 96 $\times$ 96 using a commercial face SDK. See Table~\ref{Tab:Params} for the parameters used in generating the near-duplicates. For the node embedding module, we used a training set of 1,500 IPTs generated from 7,500 face images belonging to 123 subjects, and a validation set of 600 IPTs generated from 3,000 face images belonging to 46 subjects. The number of labeled spanning trees with $n$ nodes is $n^{(n-2)}$ using Cayley's formula~\cite{cayley}. We selected six configurations as depicted in Fig.~\ref{Fig:Expt2_IPT} for training the graph neural networks, as these configurations cover maximum breadth and depth values possible for an IPT with five nodes.

\begin{figure}[h]
\centering
\subfloat[]
{
    \includegraphics[scale=0.35]{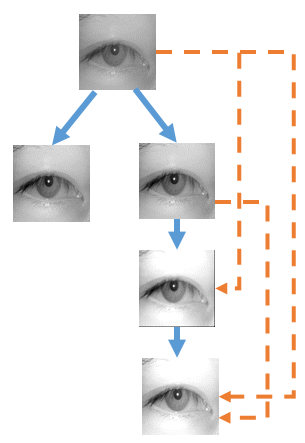}
    
} \hfill
\subfloat[]
{
    \includegraphics[scale=0.35]{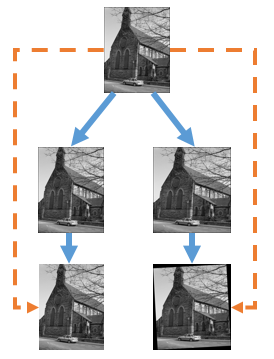}
}

\caption{IPT configuration of iris and natural scene images used for evaluation in Experiment 2.}
\label{Fig:Iris}
\end{figure}

\noindent 1) Firstly, we conducted experiments in two scenarios. In Scenario 1, we trained and tested on face images from the LFW dataset. In this scenario, the test set comprised 900 IPTs involving 4,500 face images corresponding to 75 subjects disjoint from the training and validation sets, and were evaluated separately for photometric and geometric transformations. The test IPT configurations were similar to Fig.~\ref{Fig:Expt2_IPT}. In Scenario 2, we trained the graph neural network, \textit{i.e.}, the node embedding module on face images but tested on images from the Uncompressed Color Image Database (UCID)~\cite{UCID} depicting natural scene and generic objects. We used 50 images as used in~\cite{Dias_12_MST} and applied photometric and geometric transformations to simulate 50 (number of original images)$\times$5 (number of images in each IPT)$\times$6 (number of IPT configurations)$\times$2 (photometric and geometric transformations) = 3,000 test images. See Fig.~\ref{Fig:Iris}(b). In this experiment, we analyzed the robustness of the node embedding module in handle different training and testing dataset distributions.
 
\noindent 2) Next, we performed an experiment to evaluate the performance of the proposed method in the context of \textbf{unseen transformations}. We used Photoshop to manually edit face images from the LFW dataset resulting in a set of 175 near-duplicates. We generated the near-duplicates corresponding to 35 IPTs having 4 different configurations, and each configuration has 5 nodes. We used the same protocol as followed in~\cite{TBIOM_20}. We used the Curve, Hue/Saturation, Channel Mixer, Brightness, Vibrance adjustment options and blur filters for generating the test set of Photoshopped images. We also used an Autoencoder~\cite{AEC} and BeautyGlow~\cite{BeautyGlow} (a GAN-based makeup style transfer) to generate 171 near-duplicates with deep learning-based alterations using the same protocol as observed in~\cite{TBIOM_20}. For the deep learning-based transformations, we focused on root identification accuracy as the purpose of these image editing schemes is to introduce visually imperceptible modifications while faithfully mimicking the original image.   

\noindent 3) We performed another experiment where we assessed the generalizability of the proposed method in terms of \textbf{unseen modalities} and \textbf{unseen configurations}. 
For this evaluation, we tested on 6,000 near-infrared iris images from the CASIA Iris V2 Device 2 dataset~\cite{CasV2} corresponding to 30 subjects, resulting in 1,200 IPTs, where each IPT contained 5 images (see Fig.~\ref{Fig:Iris}(a)).

\noindent 4) Finally, we conducted an experiment to evaluate whether the node embedding module can handle \textbf{unseen number of nodes}. The graph neural network was trained using IPT configurations comprising 5 nodes, but we tested it on a publicly available Near-duplicate Face Images (NDFI) \textendash  Set I~\cite{Ross_19} dataset comprised 1,229 IPTs, where each IPT consisted of 10 nodes. Therefore, the test set for this experiment comprised 12,290 images.

\subsection{Expt. 3: Bias evaluation}
\label{Expt3}
In this experiment, we evaluated the robustness of the proposed scheme in the context of demographic variations. Due to the biometric utility of the face images, it is imperative to determine whether the proposed method can handle image editing schemes across different ethnicities or adversely impact a particular demographic group. We used 1,780 images from 178 subjects in the UNCW MORPH dataset~\cite{UNCW} following a 10-node IPT configuration as seen in Fig.~\ref{Fig:IPFTest} (IPT 5). We used images from 41 Asian subjects, 33 Black (African) subjects, 38 Hispanic subjects, 29 Indian subjects and 37 White (European) subjects.

\subsection{Expt. 4: IPF reconstruction}
\label{Expt4}

\begin{figure}[h]
\centering
    \includegraphics[scale=.7]{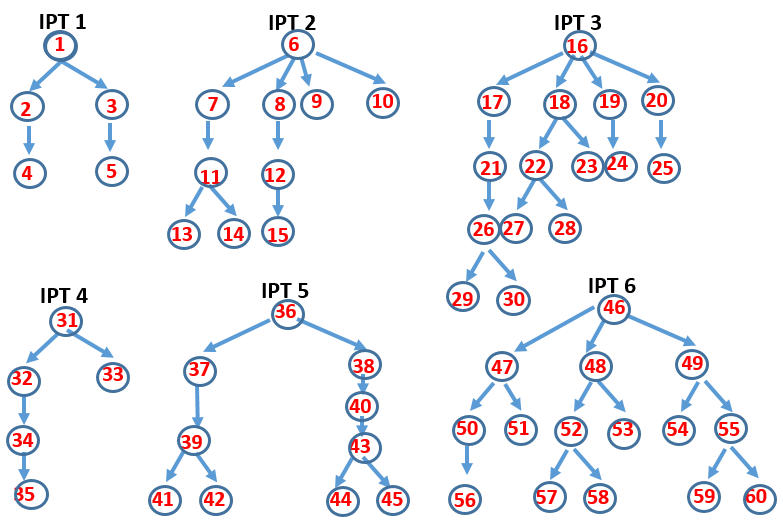} 
     
\caption{Illustration of the image phylogeny forest structures used in Experiment 4. Each IPF comprised three IPTs, with 5 nodes (IPT 1 and IPT 4) or 10 nodes (IPT 2 and IPT 5) or 15 nodes (IPT 3 and IPT 6). Test configurations differ from training configurations indicating variations in density and structures. Evaluation includes both immediate and ancestral links.}
\label{Fig:IPFTest}
\end{figure}
In this experiment, we used face images from the WVU Multimodal Release I dataset~\cite{WVU}. We used only those subjects whose images have been acquired using two different sensors: Sony EVI- D30 and Sony EVI- D39. Therefore, we used images from 49 subjects. We randomly selected 3 sample images from each of the 49 subjects. Next we subjected the images to seven transformations (four photometric and three geometric as described in Table~\ref{Tab:Params}). We used the six configurations depicted in Fig.~\ref{Fig:IPFTest} to generate 294 near-duplicate sets (49 subjects $\times$ 3 IPT configurations in each IPF $\times$ 2 sensors), such that each IPF contained three IPTs in total - the first IPT with 5 nodes, the second IPT with 10 nodes and the third IPT with 15 nodes, resulting in a total of 2,940 images (49 IPFs $\times$ 30 images in each IPF). Examples of some near-duplicates used in IPF reconstruction are presented in Fig.~\ref{Fig:Objective}. The test set contained variations in \textbf{sensors} (two sensors), \textbf{subjects} (11 females and 38 males) and \textbf{acquisition settings} (indoor and outdoor). The global bandwidth used in conventional spectral clustering was computed using the standard deviation of the respective features (face descriptors, PRNU and pixel intensities). We used three IPFs to compute the parameter for locally-scaled spectral clustering ($\eta$ = 0.7, see Algo.~\ref{Alg:LC-SC}). We report the results in terms of i) performance of the locally-scaled spectral clustering algorithm (mean number of clusters and clustering accuracy) and ii) root identification accuracy and IPF reconstruction accuracy.

\subsection{Baseline}
\label{Baseline}
We compared locally-scaled clustering with conventional spectral clustering algorithm~\cite{SpecClus_16_Dias}. We compared the proposed IPT reconstruction with Gaussian RBF and Chebyshev polynomials basis functions approach~\cite{TBIOM_20}, and the Oriented Kruskal algorithm, which has been used for reconstructing IPTs after the autoencoder-based method in~\cite{Bestagini_20_ICASSP}. 
The transformation aware embedding-based method~\cite{Bharati_20_TE} has been designed for image provenance analysis which allows multiple donor images to create a composite image. However, our method relies on a single image processing pipeline. Moreover, their codes are not available currently~\cite{Bharati_20_TE} so we could not use their method for comparison. Also, the autoencoder-based image phylogeny~\cite{Bestagini_20_ICASSP} has been designed for specific adjacency matrices (powers of 2), and could not be used for comparison with current work that offers flexibility for arbitrary number of images.

\begin{table*}[t]
\centering
\caption{Root identification and IPT reconstruction accuracy for both photometric and geometric transformations. The values to the left of the forward slash indicate Scenario 1 (trained on face images and tested on face images) and the values to the right indicate Scenario 2 (trained on face images but tested on images depicting natural scenes).}
\label{Tab:ResExpt2_NEW}
\scalebox{0.9}{
\begin{tabular}{|l|ll|ll|}
\hline
\multirow{2}{*}{\begin{tabular}[c]{@{}l@{}}IPT \\ configuration\end{tabular}} & \multicolumn{2}{c|}{Photometric transformations}                                                                                                          & \multicolumn{2}{c|}{Geometric transformations}                                                                                                            \\ \cline{2-5}
                                                                              & \begin{tabular}[c]{@{}l@{}}Root identification accuracy (\%)\end{tabular} & \begin{tabular}[c]{@{}l@{}}IPT reconstruction accuracy (\%)\end{tabular} & \begin{tabular}[c]{@{}l@{}}Root identification accuracy (\%)\end{tabular} & \begin{tabular}[c]{@{}l@{}}IPT reconstruction accuracy (\%)\end{tabular} \\ \hline
IPT A                                                                         & 90.0 / 88.0                                                                        & 94.0 / 91.33                                                                      & 86.67 / 90.0                                                                       & 91.67 / 87.33                                                                      \\
IPT B                                                                        & 69.33 / 48.0                                                                       & 78.80 / 74.80                                                                      & 46.67 / 58.0                                                                       & 74.47 / 76.0                                                                      \\
IPT C                                                                       & 86.0 / 76.0                                                                        & 96.0 / 89.67                                                                       & 74.67 / 72.0                                                                       & 93.78 / 88.0                                                                      \\
IPT D                                                                        & 98.67 / 98.0                                                                       & 97.17 / 100.0                                                                      & 95.33 / 96.0                                                                       & 95.33 / 100                                                                       \\
IPT E                                                                        & 94.0 / 92.0                                                                        & 97.73 / 93.60                                                                      & 93.33 / 86.0                                                                       & 96.27 / 94.0                                                                      \\
IPT F                                                                        & 72.67 / 46.0                                                                       & 80.15 / 76.22                                                                      & 42.67 / 50.0                                                                       & 72.37 / 74.22                                                                      \\ \hline
Average                                                                       & 85.11 / 74.67                                                                       & 90.64 / 87.60                                                                       & 73.22 / 75.33                                                                       & 87.31 / 86.72   \\ \hline                                                                  
\end{tabular}}
\end{table*}

\begin{table}[h]
\centering
\caption{Evaluation of IPT reconstruction in the context of unseen transformations, unseen modalities and configurations, and unseen number of nodes.}
\label{Tab:ResExptII_unseen}
\scalebox{0.93}{
\begin{tabular}{|l|ll|}
\hline
Experimental settings                                                                   & \begin{tabular}[c]{@{}l@{}}Root identification\\ accuracy (\%)\end{tabular} & \begin{tabular}[c|]{@{}l@{}}IPT reconstruction\\ accuracy (\%)\end{tabular} \\  \hline \hline
\textit{Unseen transformations}                                                         & 82.86                                                                       & 92.95                                                                      \\ \hline
\textit{\begin{tabular}[c]{@{}l@{}}Unseen modalities and\\ configurations\end{tabular}} & 85.92                                                                       & 90.85  
\\ \hline                                                                                                             
\textit{\begin{tabular}[c]{@{}l@{}}Unseen number of nodes\end{tabular}} & 90.97                                                                      & 61.02 \\ \hline
\end{tabular}}
\end{table}

\section{Findings and Analysis}
\label{Res}

\subsection{Results for Expt. 1: Locally-scaled clustering}
\label{ResExpt1}
\begin{figure}[]
\centering
\subfloat[]
{
    \includegraphics[scale=0.3]{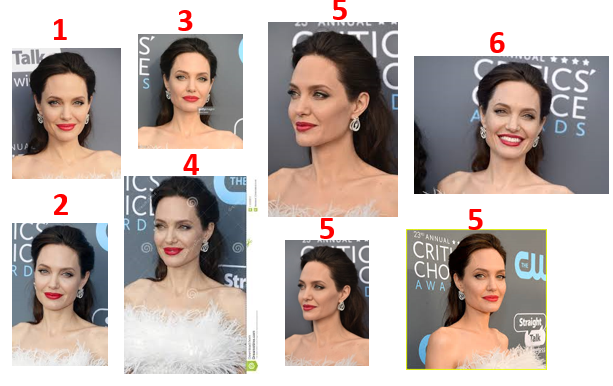}
    
}
\subfloat[]
{
    \includegraphics[scale=0.3]{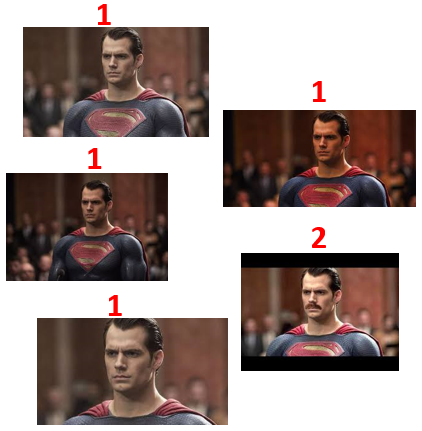}
}

\caption{Locally-scaled spectral clustering performance for near-duplicates downloaded from the \textbf{Internet}. The numbers (in red) indicate the cluster identifier for an image. (Left): Six clusters (IPTs) are identified; (Right): Two clusters (IPTs) are identified. Results are for visual inspection only, no ground truth is available.}
\label{Fig:SC_Internet}
\end{figure}

\noindent 1) We evaluated locally-scaled spectral clustering on near-duplicates downloaded from the Internet and images generated using deep learning transformations. Fig.~\ref{Fig:SC_Internet}(a) indicates that the images were clustered into six distinct IPTs (the number above the image indicates the cluster number it was assigned to). Although there is no ground truth associated with the images, we performed a qualitative analysis in this case. Some images appear to be modified with digital watermarking (such as `3' and `4'), and were correctly assigned to distinct clusters. Three images were assigned to IPT `5', although visual inspection reveals that two out of three images appear variants of each other, while the third image may have been a singleton and should have been assigned to a separate cluster. Results in Fig.~\ref{Fig:SC_Internet}(b) indicate qualitatively, correct assignment to two clusters.

\begin{figure}
\centering
    \includegraphics[scale=.3]{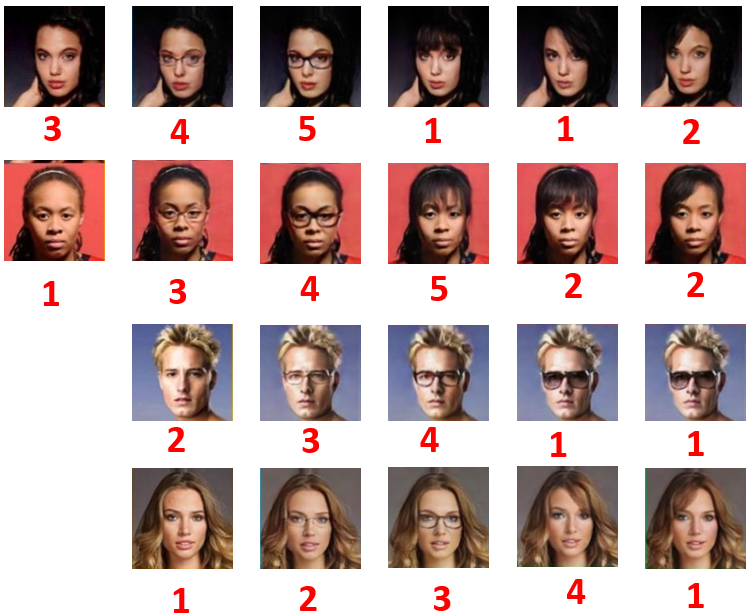} 
     
\caption{Locally-scaled spectral clustering performance for near-duplicates generated using \textbf{deep learning-based transformations}~\cite{AttGAN}. The numbers (in red) indicate the cluster identifier for an image. The proposed method was able to distinguish between attribute changes resulting in distinct clusters.}
\label{Fig:SC_GAN}
\end{figure}

\noindent 2) For the near-duplicates generated using deep learning-based method, each image was generated by modifying attributes such as adding hair bangs or adding glasses. Further modifications included making the shades darker or changing the direction of the hair bangs. The modifications were performed mostly on the original image, so we expected disjoint IPTs, as corroborated by the outputs in a majority of the cases in Fig.~\ref{Fig:SC_GAN}. 

\subsection{Results for Expt. 2: IPT reconstruction}
\label{ResExpt2}
\noindent 1) We evaluated the performance of the node embedding and link prediction module in terms of root identification accuracy and IPT reconstruction accuracy in two scenarios (see \ref{Expt2}). In Scenario 1, where both training and testing sets comprised face images, the proposed method achieved a root identification accuracy of 85.11\% in the context of photometrically modified images, and 73.22\% in the context of geometrically modified images, averaged across the six configurations. In terms of IPT reconstruction accuracy, it achieved 90.64\% in the context of photometrically modified images, and 87.31\% in the context of geometrically modified images, averaged across the six configurations.

In Scenario 2, where training was performed using face images but testing was conducted on images containing natural scenes, the proposed method achieved a root identification accuracy of 74.67\% in the context of photometrically modified images, and 75.33\% in the context of geometrically modified images, averaged across the six configurations. In terms of IPT reconstruction accuracy, the proposed method achieved 87.60\% in the context of photometrically modified images, and 86.72\% in the context of geometrically modified images, averaged across the six configurations. See Table~\ref{Tab:ResExpt2_NEW}. Results indicate that some configurations particularly II and VI were very difficult to reconstruct both for photometric and geometric transformations, indicating the difficulty in reconstructing deeper and unbalanced trees. Results further indicate that the proposed node embedding and link prediction modules were adept in handling not only different classes of transformations (photometric and geometric), but also different types of images (biometric and generic images).

\noindent 2) In the context of \textit{unseen transformations}, the proposed method achieved a root identification accuracy of 82.86\%, and an IPT reconstruction accuracy of 92.95\% averaged across four IPT configurations (see first row of Table~\ref{Tab:ResExptII_unseen}) in the context of Photoshopped images. The method achieved a root identification accuracy of 87.5\% for near-duplicates generated using Autoencoder and a root identification accuracy of 92.5\% for near-duplicates generated using the BeautyGlow network.    

\noindent 3) In the context of \textit{unseen modalities and configurations}, the proposed method achieved a root identification accuracy of 85.92\%, and an IPT reconstruction accuracy of 90.85\% in the case of iris images (see second row of Table~\ref{Tab:ResExptII_unseen}). 

\noindent 4) In the context of \textit{unseen number of nodes}, the proposed method achieved a root identification accuracy of 90.97\%, and an IPT reconstruction accuracy of 61.02\% (see third row of Table~\ref{Tab:ResExptII_unseen}). The best performing method in~\cite{Ross_19} reported a root identification accuracy of 89.91\% at Rank 3, and an IPT reconstruction accuracy of 70.61\%, assuming that the the root node is known \textit{apriori}. In contrast, we report only Rank 1 root identification accuracy. 

\begin{table}[h]
\centering
\caption{Number of clusters (mean and s.d.) produced during IPF reconstruction by conventional spectral clustering and locally-scaled spectral clustering (proposed). Lower values are desirable.}
\label{Tab:SC}
\scalebox{0.95}{
\begin{tabular}{|l|ll|}
\hline
\multirow{2}{*}{Number of nodes} & \multicolumn{2}{l|}{\begin{tabular}[c]{@{}l@{}}Number of clusters produced during IPF\\ reconstruction (mean $\pm$ s.d.)\end{tabular}} \\ \cline{2-3}
                                 & Spectral clustering              & \begin{tabular}[c]{@{}l@{}}Locally-scaled spectral\\ clustering (Proposed)\end{tabular}              \\ \hline \hline
5                                & 2.31 $\pm$ 1.04                        & \textbf{1.70 $\pm$ 0.54                                                                                           } \\
10                               & 1.98 $\pm$ 0.21                        & \textbf{1.52 $\pm$ 0.32                                                                                           } \\
15                               & 2.35 $\pm$ 0.80                        & \textbf{1.84 $\pm$ 0.53                                                                                          } \\ \hline 

\end{tabular}}
\end{table}

\subsection{Results for Expt. 3: Bias evaluation}
\label{ResExpt3}
{In the context of demographic variations, we reported the results for five different demographic groups (Asian, Black, Hispanic, Indian and White) in terms of root identification accuracy at Rank 2 and IPT reconstruction accuracy. The reason for using Rank 2 is that we observed that Rank 2 accuracy is $\sim$100\% for both Experiment 1 and Experiment 2, \textit{i.e.,} we are guaranteed to find the correct root node at Rank 2. Therefore, we want to focus on the performance variation across demographic groups by benchmarking it using the \textit{best} use-case of our method.} The top-$k$ root nodes can be identified as described in Sec.~\ref{IPT}.  We achieved 100\% root identification accuracy at Rank 2 across all five demographic groups. We achieved an IPT reconstruction accuracy of 71.4\% for Asian subjects, 68.6\% for Black subjects, 66.1\% for Hispanic subjects, 71.2\% for Indian subjects and 71.8\% for White subjects. The performance indicates strong robustness of the proposed method across different ethnic groups, even though the graph neural network is trained primarily on White subjects. This demonstrates that the method learns the global and local relationship irrespective of the scene content.     

\subsection{Results for Expt. 4: IPF reconstruction}
\label{ResExpt4}
\begin{figure}[]
\centering
    \includegraphics[scale=0.38]{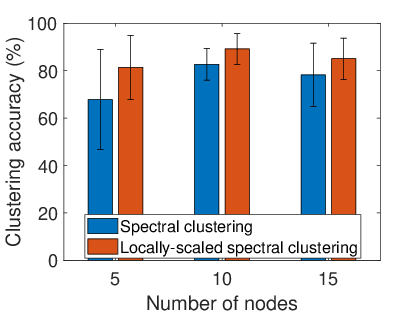}   
\caption{Variation in clustering accuracy versus the number of nodes for the conventional spectral clustering (blue) and the locally-scaled spectral clustering (orange) methods. The proposed method consistently resulted in higher means and lower standard deviations outperforming the conventional spectral clustering algorithm.}
\label{Fig:Errorplot_SC}
\end{figure}

In the context of IPF reconstruction, we first used locally-scaled spectral clustering for identifying number of IPTs, and then used node embedding and link prediction module to construct each IPT within the IPF. Table~\ref{Tab:SC} reports the number of clusters produced by conventional spectral clustering and the the locally-scaled spectral clustering. Fig.~\ref{Fig:Errorplot_SC} indicates the clustering accuracy (\textit{i.e.} the proportion of images correctly assigned to the respective clusters) for both conventional and proposed spectral clustering methods. Each IPF contained three clusters (IPTs), and each IPT comprised either 5 or 10 or 15 nodes. Results indicate that conventional clustering produced much higher number of clusters than the desired output, resulting in erroneous assignment of images. On the other hand, the proposed method performed well irrespective of the number of nodes (images). 

\begin{figure}[h]
\centering
\subfloat[]
{
    \includegraphics[scale=0.38]{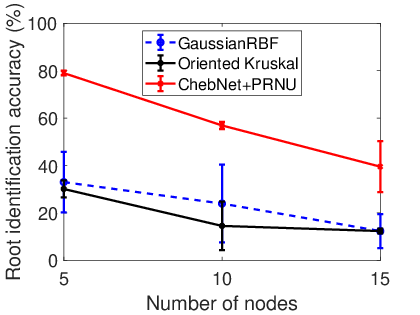}
    
}\\
\subfloat[]
{
    \includegraphics[scale=0.38]{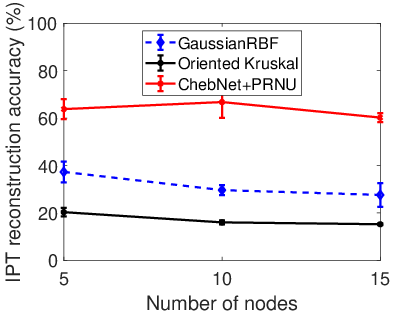}
}

\caption{Variation in root identification and IPT reconstruction accuracy versus the number of nodes.}
\label{Fig:Errorplots}
\end{figure}

We reported the root identification accuracy (the root is identified for each IPT within IPF) and IPT reconstruction accuracy for each IPT within the IPF. We also reported the average root identification and IPT reconstruction accuracies, which measures the overall IPF reconstruction performance. The proposed ChebNet-based node embedding and PRNU-based link prediction outperformed the two baselines employing Oriented Kruskal and Gaussian Radial Basis Functions (RBF)\footnote{We also compared with Chebyshev basis functions, but Gaussian RBF outperformed Chebyshev polynomials in this experiment. So we reported the results pertaining to Gaussian RBF only for the sake of brevity.} basis function (see Table~\ref{Tab:ResExpt4_GCN}) in all the cases by a significant margin. Results indicate that the proposed method outperformed Oriented Kruskal by 28.01\% and Gaussian RBF by 23.91\% in terms of root identification accuracy; the proposed method outperformed Oriented Kruskal by 42.25\% and Gaussian RBF by 27.94\% in terms of IPT reconstruction accuracy. Error plots indicating mean and standard deviations of the root identification and reconstruction accuracy are illustrated as a function of the variation in the number of nodes (images) in Fig.~\ref{Fig:Errorplots}. Results indicate that the proposed method outperformed Gaussian RBF and Oriented Kruskal-based methods.

\begin{table*}[t]
\centering
\caption{Evaluation of node embedding (ChebNet) and link prediction (PRNU) for each IPT configuration used in the IPF in terms of root identification and reconstruction accuracy. Results indicate that the proposed method (bolded) significantly outperformed state-of-the-art baselines in all the cases.}
\label{Tab:ResExpt4_GCN}
\scalebox{0.98}{
\begin{tabular}{|l|lll|lll|}
\hline
\multirow{2}{*}{IPT configuration} & \multicolumn{3}{l}{Root identification accuracy (\%)}                                                        & \multicolumn{3}{|l|}{IPT reconstruction accuracy (\%)}                                                         \\ \cline{2-7}
                                   & Oriented Kruskal & \begin{tabular}[c]{@{}l@{}}Basis functions\\ (Gaussian RBF)\end{tabular} & ChebNet+PRNU       & Oriented Kruskal & \begin{tabular}[c]{@{}l@{}}Basis functions\\ (Gaussian RBF)\end{tabular} & ChebNet+PRNU       \\ \hline \hline
IPT 1 (5 nodes)                    & 32.61            & 23.91                                                                    & \textbf{78.26} & 21.56            & 34.17                                                                    & \textbf{66.78} \\
IPT 2 (10 nodes)                  & 7.24             & 12.32                                                                    & \textbf{57.97} & 15.38            & 28.09                                                                    & \textbf{71.48} \\
IPT 3 (15 nodes)                 & 13.04            & 7.24                                                                     & \textbf{47.10} & 14.96            & 24.01                                                                    & \textbf{61.55} \\
IPT 4 (5 nodes)                   & 27.53            & 42.03                                                                    & \textbf{79.71} & 19.02            & 40.37                                                                    & \textbf{60.76} \\
IPT 5 (10 nodes)                   & 21.74            & 35.50                                                                    & \textbf{55.79} & 16.59            & 31.09                                                                    & \textbf{62.03} \\
IPT 6 (15 nodes)                  & 11.59            & 17.39                                                                    & \textbf{31.88} & 15.43            & 31.11                                                                    & \textbf{58.86} \\ \hline
Average                            & 18.96          & 23.06                                                                    & \textbf{46.97} &   17.16         & 31.47 & \textbf{59.41} \\ \hline
\end{tabular}}
\end{table*}

The \textbf{main findings} from the experiments are as follows.
\noindent 1. Locally-scaled spectral clustering can address the multi-scale issue encountered by conventional spectral clustering. This is particularly relevant in IPF reconstruction where the number of nodes (near-duplicate images) belonging to each IPT is not known \textit{apriori} (see Table~\ref{Tab:SC}). Experiments indicate that the proposed method performed robustly when the number of nodes in an IPT was varied by a factor of two and three in an IPF, and appropriated well to near-duplicates downloaded from the Internet or generated using deep learning tools.\\ 
\noindent 2. Graph theoretic approach integrated with sensor pattern noise significantly outperformed existing methods in constructing image phylogeny tree. Experiments indicate that the proposed method performed very well for both photometrically and geometrically modified face as well as natural scene images (see Table~\ref{Tab:ResExpt2_NEW}). The proposed method also generalized well across unseen transformations, unseen modalities, unseen configurations and unseen number of nodes. See Table~\ref{Tab:ResExptII_unseen}. \\
\noindent 3. Locally-scaled spectral clustering used together with node embedding and link prediction outperformed existing methods in the context of image phylogeny forest reconstruction, and offered substantial improvement, with upto $\sim$ 28\% in terms of root identification accuracy and upto $\sim$ 42\% in terms of IPF reconstruction accuracy. See Table~\ref{Tab:ResExpt4_GCN}.

\subsection{Ablation}
\label{Sec:Ablation}
{ \textbf{Comparison between different node embedding module architectures.} We compared the performance of three node embedding techniques used in this work \textit{viz.}, GCN, ChebNet and HGNN, and selected the network yielding the highest depth classification accuracy on the test set. The node labels (depth values) from the best performing model in the first stage were further fed to the link prediction module to reconstruct the IPT. Results indicate that ChebNet with Chebyshev polynomial of degree three outperformed GCN and HGNN by a considerable margin ($ \approx 20\%$) in depth label prediction, and was, therefore, selected as the \textit{best} node embedding technique out of the three models used in this work. We subsequently used the depth labels provided by the ChebNet in the link prediction module. The reason for ChebNet to perform better than the remaining two methods could be attributed to the fact that ChebNet covers the full spectrum profile compared to GCN, as indicated in~\mbox{\cite{ChebvsGCN}}. HGNN builds on top of the linear approximation used in GCN, and follows similar performance.}

\begin{figure*}[t]
\centering
    \includegraphics[width=0.6\textwidth]{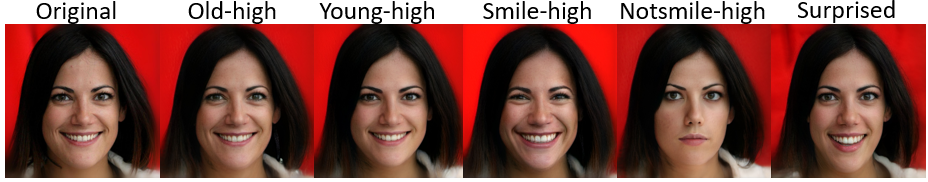} 
\caption{{Examples of images used in \textbf{training} the graph neural network from the CelebAHQ-FM dataset. Note the training set consisted of total six images for a single individual (one original image plus five edited images). The edits are: \textit{notsmile-high, smile-high, old-high, young-high}, and \textit{surprised}.} }
\label{Fig:STRNtrain}
\end{figure*}

\begin{figure*}[t]
\centering
    \includegraphics[width=0.6\textwidth]{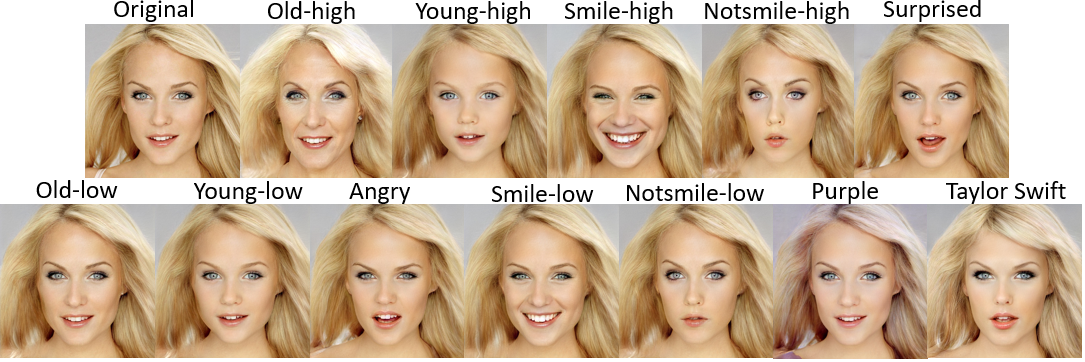} 
\caption{{Examples of images used in \textbf{testing} the graph neural network from the CelebAHQ-FM dataset for node embedding. Note the test set consisted of total thirteen images for a single individual (one original image plus 12 edited images) for disjoint subjects. The edits are: \textit{notsmile-high, notsmile-low, smile-high, smile-low, old-high, old-low, young-high, young-low, surprised, angry, purple-hair}, and \textit{Taylor Swift}. Note there are overlap of 5 edits from the training set and the remaining 7 edits are ``unseen''. These include editing attributes to different degrees (low vs high).}}
\label{Fig:STRNtest}
\end{figure*}

\begin{table}[h]
\centering
\caption{{Comparison between pixel, PRNU and convolutional features on the root identification accuracy on the \textit{CelebAHQ-FM} dataset. The graph neural network requires features of the nodes ($\mathbf{X}$) and the adjacency matrix ($\mathbf{A}$). We utilize two types of convolutional features in our work, namely, \textbf{Inception-ResNet-V2} (IRNet) and \textbf{ResNet100-ArcFace} (ArcFace). We report accuracy for each convolutional feature using forward slash notation.} }
\begin{tabular}{ccc} \hline
\multicolumn{1}{c}{$\mathbf{X}$} & $\mathbf{A}$                   & \begin{tabular}[c]{@{}c@{}}Root identification\\ accuracy (\%)\end{tabular} \\ \hline
Pixel                 & PRNU                & 99.32                                                                  \\
Pixel                 & IRNet/ArcFace & 46.23/0.0                                                                  \\
IRNet/ArcFace   & PRNU                & 99.32/88.36   \\ \hline                                                              
\end{tabular}
\label{Tab:Rev2_q1}
\end{table}

\begin{figure*}[t]
\centering
    \includegraphics[width=0.7\textwidth]{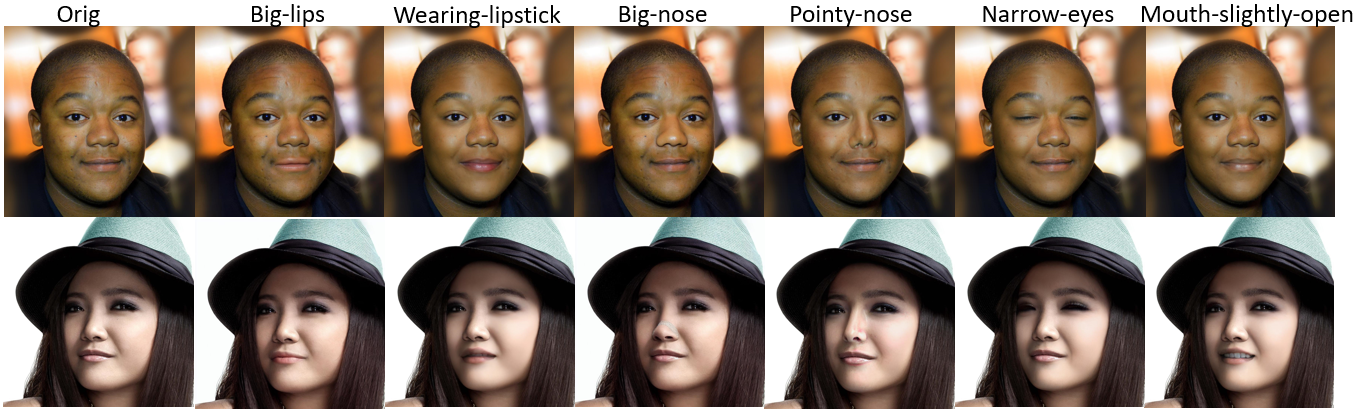} 
\caption{{Examples of images used for testing the generalizability of the proposed method on images generated using MaskFaceGAN and ControlNet-inpainting using diffusion models. The test set consists of seven images for a single individual (one original image plus six edited images). The edits are: \textit{Big-lips, Wearing-lipstick, Big-nose, Pointy-nose, Narrow-eyes}, and \textit{Mouth-slightly-open}. Note that the neural networks and the editing operations are significantly different from the CelebAHQ-FM dataset.}}
\label{Fig:Diffmodel}
\end{figure*}

{ \textbf{Performance on near-duplicates from facial edits via generative models.} We further evaluated our method on transformations by state-of-the-art generative models. Firstly, we used the CelebAHQ-FM dataset from the Comprehensive Dataset of Face Manipulations for Development and Evaluation of Forensic Tools~\mbox{\cite{STRN}}. The original face images from the CelebAHQ dataset were altered using \textit{pivotal tuning}~\mbox{\cite{PTI}} for performing realistic facial edits by manipulating the latent space of a StyleGAN model. The training set consists of \textit{notsmile-high, smile-high, old-high, young-high}, and \textit{surprised} edits. Note \textit{high} corresponds to the degree or the intensity of variation for that particular attribute. The test set consists of \textit{notsmile-high, notsmile-low, smile-high, smile-low, old-high, old-low, young-high, young-low, surprised, angry, purple-hair}, and \textit{Taylor Swift} edits. Both training and test sets are subject disjoint and contain the original image. Note that the test set contains seven \textit{unseen} attributes not present in the training set. Refer to examples from the training set in Fig.~\mbox{\ref{Fig:STRNtrain}} and from the test set in Fig.~\mbox{\ref{Fig:STRNtest}}. Our method achieves root identification accuracy of 99.33\% on this challenging dataset. Secondly, we simulated additional facial edits not present in the CelebAHQ-FM dataset by using the latest \textit{MaskFaceGAN}~\mbox{\cite{maskfacegan}} and \textit{ControlNet-inpainting using diffusion model}~\mbox{\cite{cnip}} We used $\sim$100 subjects from the CelebAHQ dataset and applied GAN and diffusion models to produce six facial edits \textit{not present} in the CelebAHQ-FM dataset. The edits are: \textit{Big-lips, Wearing-lipstick, Big-nose, Pointy-nose, Narrow-eyes}, and \textit{Mouth-slightly-open}; see examples in Fig.~\mbox{\ref{Fig:Diffmodel}}. Some edited images had visible artifacts, so we discarded them. Finally, we had a set of 679 images from 97 subjects. We obtained each transformed image by editing the original image, so, we had 97 IPTs with original image as the \textsc{root node} and the six edited images at depth label=2, \textit{i.e.}, \textsc{leaf nodes}. We tested the \textit{generalizability} of our method by training on the CelebAHQ-FM dataset and testing on the simulated set. Note that the simulated set differed from the training set in terms of unseen facial edit operations and different neural networks used for editing. CelebAHQ-FM (training set) used StyleGAN+pivotal tuning whereas, the test set used GAN+Diffusion model. This will inevitably result in significant variations between the training and test distributions. Our method performed reasonably well with 70\% root identification accuracy and 70\% IPT reconstruction accuracy.}

{ \textbf{Impact of convolutional feature maps as inputs to graph neural network.} The graph neural network, $f(\mathbf{X}, \mathbf{A})$ accepts a pair of inputs: features of the nodes ($\mathbf{X}$) and the adjacency matrix ($\mathbf{A}$). The adjacency matrix construction also requires features. We ablated using \textit{convolutional feature maps} in conjunction with pixel intensity feature values and sensor pattern noise features. Therefore, we examined three different scenarios: (1) $\mathbf{X}\leftarrow$ pixel values, $\mathbf{A}\leftarrow$ sensor pattern noise, (2) $\mathbf{X}\leftarrow$ pixel values, $\mathbf{A}\leftarrow$ convolutional features and (3) $\mathbf{X}\leftarrow$ convolutional features, $\mathbf{A}\leftarrow$ sensor pattern noise. Pixel intensity values are simply flattened intensity values of the face images (we only used the first channel and resized them to 96$\times$96 for compactness); sensor pattern noise values are computed using the Enhanced PRNU features~\mbox{\cite{Li_TIFS_10}}, and convolutional features are computed using two networks: (a) Inception-ResNet-V2~\mbox{\cite{IRNet}} that has 824 layers, and trained on the ImageNet dataset. We use the features from the final average pooling layer corresponding to $\mathbb{R}^{1 \times 1536}$ embedding for each image. (b) ResNet100-ArcFace~\mbox{\cite{arcface}} as the face recognition model trained with ArcFace loss on the Glint360K face dataset from the InsightFace package with antelopev2 model. We used the output features of the model corresponding to $\mathbb{R}^{1 \times 512}$ embedding for each image. Refer to the results obtained in Table~\mbox{\ref{Tab:Rev2_q1}}.} 

{ \textbf{Comparison with image provenance algorithms.} For this ablation study, we compared our performance with~\mbox{\cite{Moreira_18_Provenance}}. Although their objective was different than ours, their graph construction for the near-duplicates can be used as baseline against our method. Therefore, we adapted the official code provided by the original authors~\footnote{\url{https://github.com/CVRL/Scalable_Provenance/tree/master}} and compared with the graph building framework used in their paper. The baseline method performs the following steps. (i) \textit{GCM-based dissimilarity calculation} in which they perform interest point detection on a set of near-duplicates followed by geometrically consistent matching (GCM) of the interest points. (ii) \textit{MI-based dissimilarity calculation} in which they perform homography estimation from the GCM followed by mutual information (MI) calculation from the warped and color-matched patches. (iii) \textit{Construction of provenance graph} in which they perform Kruskal's Minimal Spanning Tree algorithm using the dissimilarity matrix. Refer to~\mbox{\cite{Moreira_18_Provenance}} for details. We used the SURF descriptor and similarity transformation for homography estimation. As the dissimilarity matrix from MI was asymmetric, we used the Oriented Kruskal~\mbox{\cite{Dias_13_OB}} algorithm (classical Kruskal's algorithm needs symmetric matrix). We achieved root identification and IPT reconstruction accuracy $\sim 0.1\%$. We tried different combinations of using only GCM-dissimilarity matrix, MI-dissimilarity matrix and combined (min-max normalized) dissimilarity matrix. However, we did not observe improvement in performance. Note MI captures similarity between two images, the authors in~\mbox{\cite{Moreira_18_Provenance}} use it to determine relationship between the query and the donor images that were used to generate the composite. However, in our scenario with near-duplicate faces with exactly same pose and background, implying source and target images are perfectly registered, mutual information is not discriminative enough to deduce the relationship between the near-duplicates. Our findings reaffirm that mutual information and geometric consistent matching may work on composite images originating from different donor images but fails to distinguish between subtle edits in face images.}



\section{Conclusion and Future Work}
\label{Concl}

Image phylogeny tree generation from a set of near-duplicate images is a fundamentally challenging problem. This is further compounded when these images correspond to faces with different poses, expressions and acquisition settings. This results in multiple phylogeny trees, \textit{i.e.}, an image phylogeny forest. We use a locally-scaled spectral clustering algorithm that can robustly handle multi-scale forests. Next, we formulate the image phylogeny as a graph-based problem with near-duplicates as nodes and edges representing their relationship. We use a graph neural network to deduce the global hierarchical position of an image in the IPT, known as the depth label, followed by the use of sensor pattern noise features for predicting local links between the parent and child nodes. Thus we leverage both global and local dependencies simultaneously to construct an accurate IPT for face images. Experiments demonstrate the efficacy of the proposed method on diverse datasets. The proposed method is generalizable to images from different biometric modalities as well as natural scene images, is robust to unseen transformations and IPT configurations. It outperforms the baselines with a significant margin ($\sim$ 42\%) in terms of IPF reconstruction accuracy.

Future work will focus on integrating the two-step approach into a single framework for IPT reconstruction. We will incorporate spurious edge pruning to further improve the IPT reconstruction accuracy. 
\balance

\bibliographystyle{IEEEtran}
\bibliography{TBIOM}

\begin{IEEEbiography}[{\includegraphics[width=1in,height=1.25in,clip]{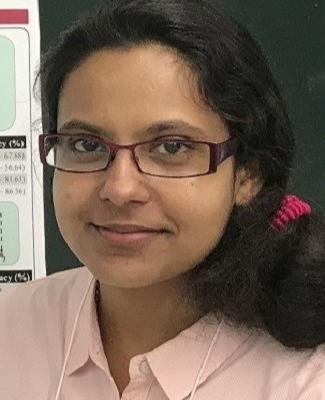}}]{Sudipta Banerjee} is a Research Assistant Professor in the Department of Computer Science at New York University, US. She received her Bachelor in Technology from WBUT, India in 2011, Masters' in Electronics and Telecommunication Engineering from Jadavpur University, India in 2014 and Doctorate in Computer Science from Michigan State University, US in 2020. Her research focuses on biometrics, generative modeling and image forensics.
\end{IEEEbiography}

\begin{IEEEbiography}[{\includegraphics[width=1in,height=1.25in,clip]{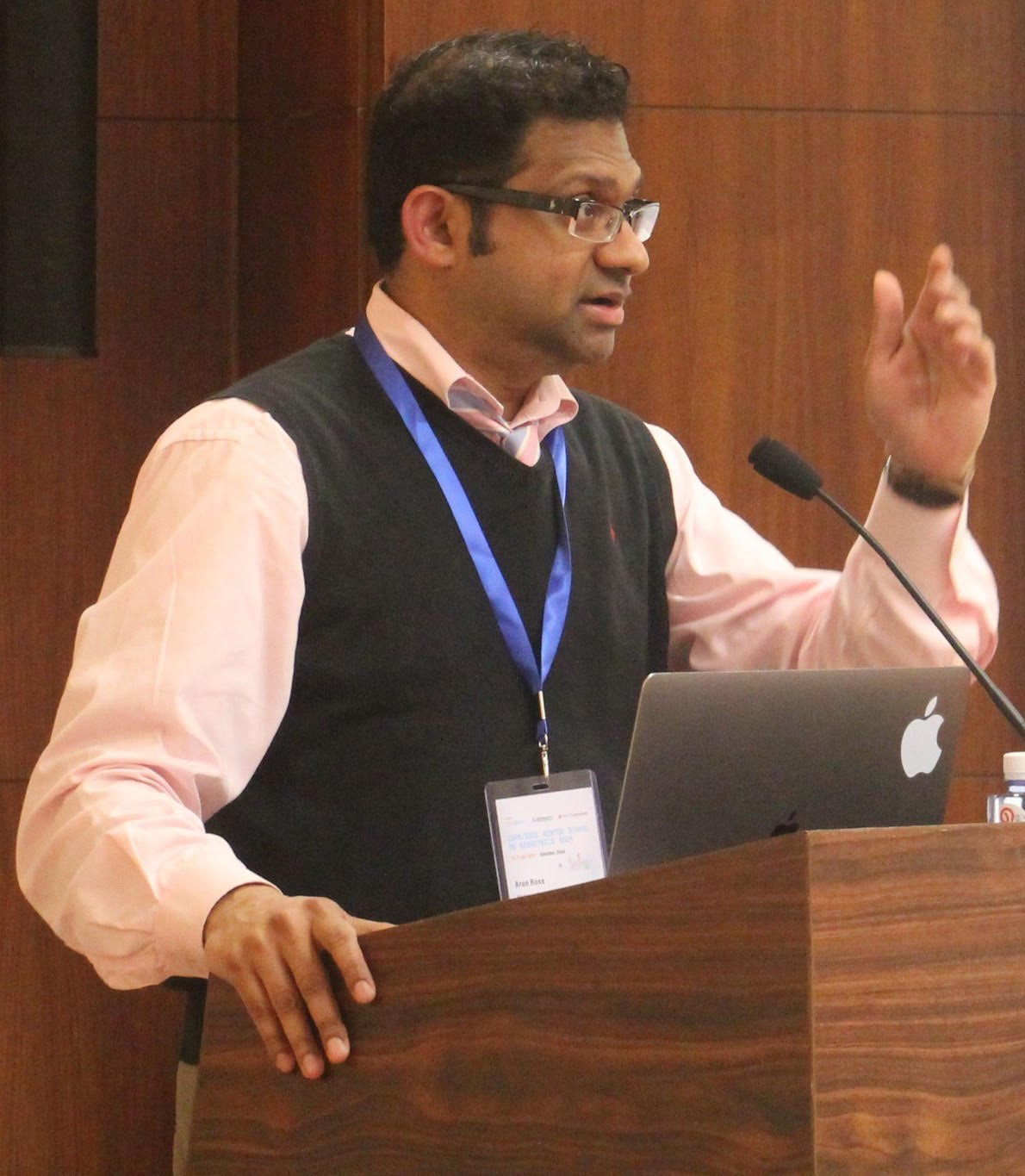}}]{Arun Ross} is the Martin J. Vanderploeg Endowed Professor in the Department of Computer Science and Engineering at Michigan State University, and the Site Director of NSF’s Center for Identification Technology Research (CITeR). He received the B.E. (Hons.) degree in Computer Science from BITS Pilani, India, and the M.S. and PhD degrees in Computer Science and Engineering from Michigan State University. Ross is a recipient of the JK Aggarwal Prize (2014) and the Young Biometrics Investigator Award (2013) from the International Association of Pattern Recognition for his contributions to the field of Pattern Recognition and Biometrics. He was designated a Kavli Fellow by the US National Academy of Sciences by virtue of his presentation at the 2006 Kavli Frontiers of Science Symposia. Ross is also a recipient of the NSF CAREER Award.
Ross has advocated for the responsible use of biometrics in multiple forums including the NATO Advanced Research Workshop on Identity and Security in Switzerland in 2018. He testified as an expert panelist in an event organized by the United Nations Counter-Terrorism Committee at the UN Headquarters in 2013. In June 2022, he testified at the US House Science, Space, and Technology Committee on the topic of Biometrics and Personal Privacy. He is a co-author of the monograph “Handbook of Multibiometrics” and the textbook “Introduction to Biometrics”.
\end{IEEEbiography}

\end{document}